\documentclass{article}

\usepackage[sglblindworkshop]{neurips_2026}

\bibpunct{[}{]}{,}{n}{}{}

\workshoptitle{Muslim in ML}

\usepackage[utf8]{inputenc}
\usepackage[T1]{fontenc}
\usepackage{hyperref}
\usepackage{url}
\usepackage{booktabs}
\usepackage{amsfonts}
\usepackage{nicefrac}
\usepackage{microtype}
\usepackage{xcolor}
\usepackage{graphicx}
\usepackage{amsmath}
\usepackage{lineno} % Defines linenomath used by neurips_2026.sty
\usepackage{caption}
\usepackage{subcaption}
\usepackage{float}
\usepackage{placeins}   % \FloatBarrier

\title{They Know Where It Is, But Not What It Says: A Bilingual Hallucination Benchmark of Hadith Recall in Open LLMs}

\author{%
  \begin{tabular}[t]{ccc}
    \textbf{Dayam Nadeem} & \textbf{Safdar Mustafa} & \textbf{Ameen Ur Rehman} \\[2pt]
    \small Jamia Hamdard & \small Jamia Hamdard & \small KFUPM \\[2pt]
    \small\texttt{dayam8696@gmail.com} & \small\texttt{safdarmustafa01@gmail.com} & \small\texttt{g202527430@kfupm.edu.sa}
  \end{tabular}
  \AND
  \hspace{-2em}
  \begin{tabular}[t]{ccc}
    \textbf{Shahab Saquib Sohail}$^{*}$ & \textbf{Mohammad Nadeem} & \textbf{Usman Naseem} \\[2pt]
    \small Jamia Hamdard & \small Aligarh Muslim University & \small Macquarie University \\[2pt]
    \small\texttt{shahabssohail@jamiahamdard.ac.in} & \small\texttt{mnadeem.cs@amu.ac.in} & \small\texttt{usman.naseem@mq.edu.au}
  \end{tabular}
}

\begin{document}

\maketitle

% ─────────────────────────────────────────────────────────────────────────────
  \begin{abstract}
  Large language models are increasingly consulted for religious knowledge,
  including the recall of specific Hadith (recorded sayings and actions of the
  Prophet Muhammad, peace be upon him). Unlike open-ended generation, Hadith
  recall is a high-stakes, verifiable task: a reference (collection, chapter,
  and Hadith number) has exactly one correct text, and any deviation from it
  constitutes a fabrication that can be mistaken for authentic scripture. This
  study introduces a bilingual (English/Arabic) benchmark that probes five open-weight,
  locally-deployable language models (Mistral-7B, Gemma3-4B, Jais-7B, ALLaM-7B,
  and Phi4-mini-3.8B) across all six canonical Sunni Hadith collections
  (\emph{Sahih al-Bukhari}, \emph{Sahih Muslim}, \emph{Sunan Abi Da'ud},
  \emph{Jami' al-Tirmidhi}, \emph{Sunan an-Nasa'i}, \emph{Sunan Ibn Majah}),
  sampling 500 Hadith per collection (3,000 per model per language, 30,000
  responses total). Each model is given only the bibliographic reference and asked to reproduce
  the exact Hadith text, with no retrieval augmentation, and is scored on
  lexical fidelity (BLEU) and semantic fidelity (multilingual cosine
  similarity), with refusals flagged separately. Across all five models and
  both languages, 0\% of 30,000 responses achieve both high lexical and high
  semantic fidelity, i.e., an exact, verifiable match to the source text.
  Arabic responses fare uniformly worse than English. These findings suggest that
  current open-weight LLMs, used without retrieval grounding, are unreliable
  and potentially dangerous sources for exact religious-text recall, and that
  deployment for Islamic scholarship should mandate retrieval-augmented
  generation over a verified corpus rather than parametric memory.  The complete code, prompts, evaluation pipeline, and scored outputs are publicly available at \url{https://github.com/dayam8696/Hadith-Hallucination-Benchmark}.
  \end{abstract}

% ─────────────────────────────────────────────────────────────────────────────
\section{Introduction}
\label{sec:intro}

Large language models (LLMs) are used, informally and increasingly, as
reference tools for religious knowledge: users ask a chatbot to quote a
specific verse, ruling, or Hadith rather than opening a physical text or a
vetted digital corpus. For most everyday knowledge tasks this convenience is
acceptable because errors are low-stakes and easily checked. Religious
scripture is different. A Hadith is a specific, canonicalized unit of
text attributed, numbered, and preserved across centuries of
chain-of-narration (\emph{isnad}) scholarship and a model that produces a
fluent, confident, but fabricated Hadith text is not merely wrong; it is
putting invented words into the mouth of the Prophet (peace be upon him) under
the appearance of authentic transmission. This is a uniquely severe form of
hallucination: the failure mode is not an unhelpful answer, but a convincing
forgery.

One might reasonably expect that a model built within the Arab world, and
trained explicitly for Arabic fluency, would recall Arabic religious text
more faithfully than a general-purpose model trained mostly on English. The
results tell the opposite story: Jais and ALLaM, both developed for
Arabic-language understanding, hallucinate at rates no better, and on
several collections worse, than models with no Arabic specialization at
all. Fluency in a language, it turns out, is not the same as having
memorized its scripture.

Despite this, no existing benchmark to date isolates
\emph{exact-recall} hallucination for Hadith text specifically, across
multiple canonical collections, in both Arabic (the original language of
transmission) and English (the language most commonly used by
non-Arabic-speaking Muslims and by LLMs deployed globally). Existing
hallucination benchmarks for LLMs largely target general factual QA,
summarization consistency, or code correctness, and multilingual
religious-text benchmarks are sparse in the open literature. This gap matters
because Hadith recall has a property that makes it an especially clean
hallucination probe: for a given reference, there is exactly one correct
answer, drawn from a fixed, well-documented corpus, so lexical and semantic
similarity to the ground truth can be measured directly, without the ambiguity
that plagues open-ended factuality evaluation.

\paragraph{Contributions.}
This study presents the first systematic bilingual benchmark of
Hadith-recall hallucination in open-weight LLMs, to the best of current
knowledge. Concretely:
\begin{itemize}
  \item This study constructs a stratified, reproducible evaluation set of
    3,000 Hadith references per language (500 randomly sampled from each of
    the six major Sunni collections), paired with ground-truth Arabic and
    English text.
  \item This study evaluates five open-weight models runnable on consumer
    hardware via Ollama including two models built specifically for Arabic
    (Jais-7B, ALLaM-7B) under a reference-only prompt with no retrieval
    augmentation, in both Arabic and English.
  \item This study proposes a two-axis scoring protocol (lexical fidelity via
    BLEU, semantic fidelity via multilingual sentence-embedding cosine
    similarity) with explicit refusal detection, and uses it to classify
    every response as \emph{correct}, \emph{partial}, \emph{theme-only},
    \emph{hallucinated}, or \emph{refused}.
  \item This study shows that across all five models and both languages, the
    \emph{correct} category is empty: 0\% of 30,000 responses simultaneously
    match the ground truth lexically and semantically, and Arabic responses are
    uniformly classified as hallucinated regardless of model or collection.
\end{itemize}
Figure~\ref{fig:pipeline} gives a complete visual walkthrough of this
pipeline end to end: how a bare bibliographic reference becomes a bilingual
prompt, how that prompt is answered by each of the five evaluated models,
and how every resulting response is filtered, scored, and finally sorted
into one of five outcomes. Rather than describing each stage in the
abstract, the figure lets the reader trace a single Hadith reference through
the entire system at a glance; each step is unpacked in turn in
Section~\ref{sec:benchmark} and Section~\ref{sec:eval}.

% ─────────────────────────────────────────────────────────────────────────────
\section{Related Work}
\label{sec:related}

\paragraph{Hallucination benchmarks.}
General-purpose hallucination and factuality benchmarks (e.g., TruthfulQA-style
probes, retrieval-free factual QA suites) typically evaluate open-domain trivia
or encyclopedic facts, where partial credit and paraphrase are acceptable.
Hadith recall differs in requiring near-verbatim fidelity to a fixed source
text, which allows translation-style lexical metrics (BLEU) to be applied
directly as a hallucination signal rather than relying solely on binary
factuality judgments.

\paragraph{Religious and low-resource-domain NLP.}
Prior work on Arabic and Islamic NLP has largely focused on classification
(authenticity/\emph{sanad} grading of Hadith, topic classification of Quranic
verses) rather than generative recall. Arabic-centric LLMs such as
Jais~\cite{jais} and ALLaM~\cite{allam} have been evaluated on general Arabic
benchmarks (MMLU-Arabic-style suites, translation) but not, to date, on
verbatim recall of a fixed religious corpus, which is the setting targeted
here.

\paragraph{Retrieval-augmented generation (RAG).}
These results are consistent with the broader RAG literature's premise that
parametric memory alone is an unreliable substitute for grounding in a verified
corpus for knowledge-intensive, low-tolerance-for-error tasks. This benchmark
is presented as a motivating case for why Islamic-scholarship LLM applications
should be retrieval-grounded rather than relying on model memorization.

% ─────────────────────────────────────────────────────────────────────────────
\section{Benchmark Construction}
\label{sec:benchmark}

Figure~\ref{fig:pipeline} lays out the complete methodology end to end,
tracing the path from a raw Hadith reference all the way to a final
classification label. It shows, in a single view, how the source corpus is
sampled, how each sampled reference becomes a bilingual prompt, how that
prompt is routed to each of the five evaluated models, and how the
resulting response is filtered for refusals, scored for lexical and
semantic fidelity, and classified. The remainder of this section walks
through each of these stages in the order the figure presents them, with
the scoring details deferred to Section~\ref{sec:eval}.

\subsection{Source Corpus}
This study uses the publicly available Hadith Dataset~\cite{hadithdataset},
released on Kaggle under a CC0 (public domain) license. The dataset provides
parallel Arabic and English text together with collection name, chapter
number and name, Hadith number, and a narrator chain-of-transmission index
for every record. After deduplication and a cleaning pass on whitespace and
encoding artifacts, the working corpus contains 34,441 Hadith spanning the six
canonical Sunni collections (\emph{kutub al-sittah}): Sahih
al-Bukhari~\cite{bukhari} (7,370), Sahih Muslim~\cite{muslim} (7,596), Sunan
Abi Da'ud~\cite{abudawud} (5,260), Jami' al-Tirmidhi~\cite{tirmidhi} (4,214),
Sunan an-Nasa'i~\cite{nasai} (5,774), and Sunan Ibn Majah~\cite{ibnmajah}
(4,227).

% We note, and flag as a limitation in Section~\ref{sec:limitations}, that the
% dataset's creator explicitly describes it as an unvalidated draft; we treat it
% as the best available open, parallel Arabic/English Hadith corpus for
% benchmarking purposes.

\subsection{Sampling}
From this corpus, a stratified random sample of 500 Hadith per collection
(3,000 total) is drawn, using a fixed random seed for reproducibility.
Sampling is per-book to guarantee balanced coverage of every collection
regardless of its size in the source corpus.

\subsection{Prompting Protocol}
\begin{figure}[t]
  \centering
  \includegraphics[width=\linewidth, height=0.85\textheight, keepaspectratio]{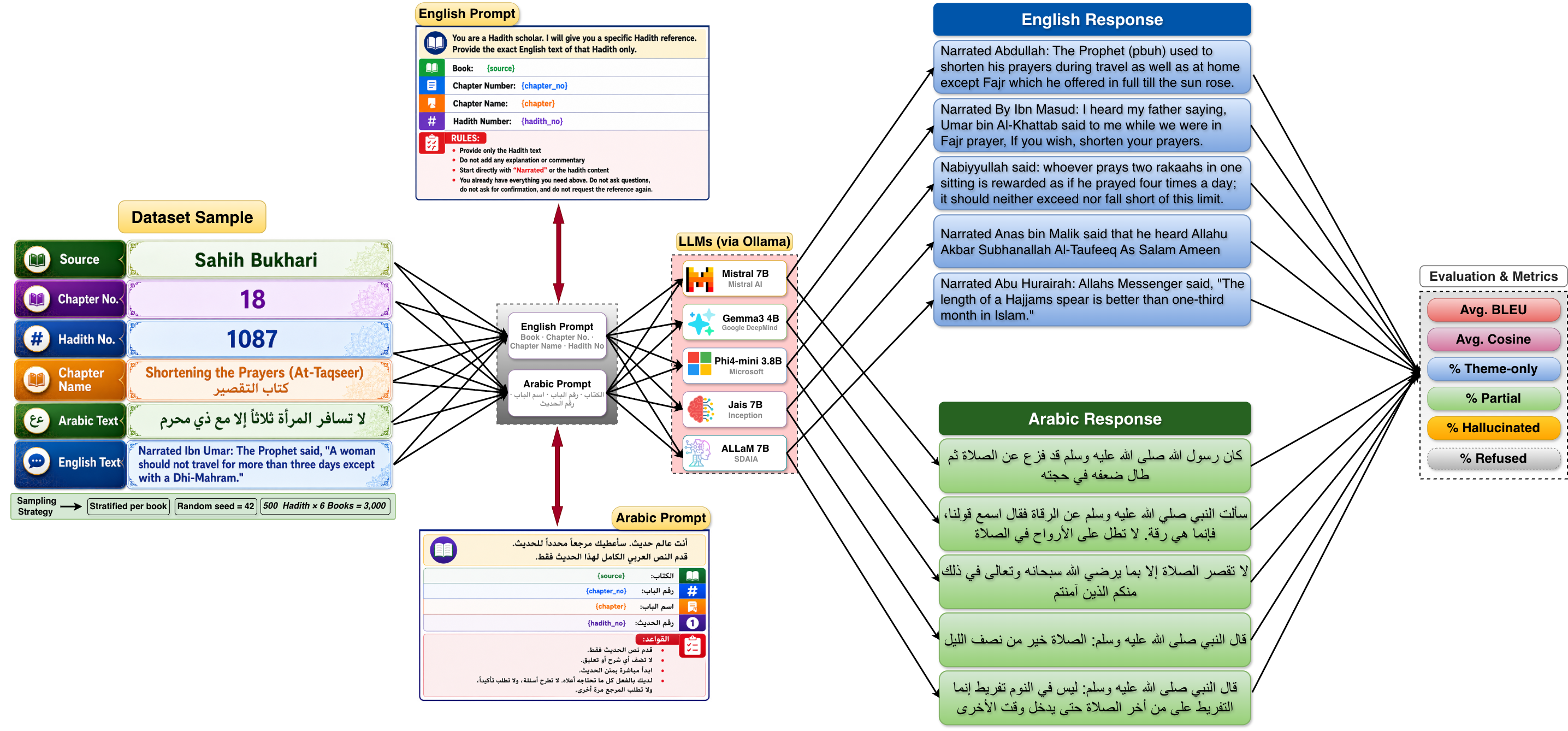}
  \caption{End-to-end evaluation pipeline: reference-only prompt
    $\rightarrow$ local LLM inference (Ollama) $\rightarrow$ model response
    $\rightarrow$ refusal filter $\rightarrow$ dual scoring (BLEU lexical
    match; multilingual cosine semantic match) $\rightarrow$ five-way
    classification (\emph{correct} / \emph{theme-only} / \emph{partial} /
    \emph{hallucinated} / \emph{refused}).}
  \label{fig:pipeline}
\end{figure}

Each model receives only the bibliographic reference collection name, chapter
number, chapter name, and Hadith number and is instructed to reproduce the
Hadith text exactly, with no explanation or commentary, and with an explicit
instruction not to ask for clarification. No retrieval context or search access
is provided: the model must answer from parametric memory alone. Prompts are
issued in two parallel forms, English and Arabic, with language-matched
instructions; the structure of both prompt variants, and how each is routed
to every evaluated model, is shown in Figure~\ref{fig:pipeline}.

% ── Methodology pipeline ────────────────────────────────────────────────────

\subsection{Models}
This study evaluates five open-weight models, all served locally via
Ollama~\cite{ollama}, chosen to span general-purpose and Arabic-specialized
instruction-tuned models runnable on consumer hardware (8--16\,GB RAM):
Mistral-7B~\cite{mistral}, Gemma3-4B~\cite{gemma3}, Phi4-mini-3.8B~\cite{phi4mini},
Jais-7B~\cite{jais} (Arabic-adapted), and ALLaM-7B~\cite{allam}
(Arabic-centric). Decoding is deterministic-leaning (low temperature,
repetition penalty) to reduce degenerate repetition and isolate content
hallucination from decoding artifacts.

\section{Evaluation Protocol}
\label{sec:eval}

Let $r$ denote a model response and $g$ the ground-truth Hadith text. Every
response is passed through the pipeline below in order: refusal detection, then
(if not refused) lexical scoring, semantic scoring, and classification.

\subsection{Refusal Detection}
\label{sec:refusal}
Responses in which the model declines to answer or requests clarification
are flagged. Let $\mathcal{P}_{\text{lang}}$ be the fixed list of refusal
phrases and $L_{\max}=150$ characters. A response is a refusal if:
\begin{equation}
  \mathrm{is\_refusal}(r)
    = \bigl[\,|r| \le L_{\max}\,\bigr]
      \wedge
      \bigl[\,\exists\,p \in \mathcal{P}_{\text{lang}} : p \in r\,\bigr].
  \label{eq:refusal}
\end{equation}
The length cutoff separates a genuine refusal from a response that hedges
briefly and then still fabricates a full Hadith, which is scored as a
hallucination rather than a refusal.

\subsection{Lexical Fidelity: BLEU}
\label{sec:bleu}
Sentence-level BLEU~\cite{bleu} is computed between the whitespace-tokenized
response $r$ and ground truth $g$:
\begin{align}
  \mathrm{BP}(r,g) &=
    \begin{cases}
      1 & \text{if } |r|_{\text{tok}} > |g|_{\text{tok}}, \\
      \exp\!\bigl(1 - |g|_{\text{tok}}/|r|_{\text{tok}}\bigr)
        & \text{otherwise,}
    \end{cases}
  \label{eq:bp} \\[4pt]
  \mathrm{BLEU}(r,g) &=
    \mathrm{BP}(r,g)\cdot
    \exp\!\Bigl(\sum_{n=1}^{4} w_n \log p_n(r,g)\Bigr),
    \quad w_n = \tfrac{1}{4}\ \forall n.
  \label{eq:bleu}
\end{align}
$n$-grams up to $n{=}4$ are used with additive smoothing. $\mathrm{BLEU}(r,g)
\in [0,1]$; a response that truly reproduces the Hadith scores near 1.0.
Smoothing is needed because unsmoothed BLEU collapses to zero whenever any
single $n$-gram order has no matches, which is common given how short Hadith
texts are relative to typical machine-translation segments.

\subsection{Semantic Fidelity: Cosine Similarity}
\label{sec:cosine}
Let $E(\cdot)$ denote the multilingual sentence encoder
\texttt{paraphrase-multilingual-MiniLM-L12-v2}~\cite{sbert}, $E(\cdot)\in\mathbb{R}^{384}$.
With $\mathbf{u}=E(r)$ and $\mathbf{v}=E(g)$:
\begin{equation}
  \mathrm{cos}(r,g)
    = \max\!\Bigl(0,\;\min\!\Bigl(1,\;
        \frac{\mathbf{u}\cdot\mathbf{v}}
             {\lVert\mathbf{u}\rVert\,\lVert\mathbf{v}\rVert}
      \Bigr)\Bigr).
  \label{eq:cosine}
\end{equation}
This captures topical relatedness even when exact wording differs, giving
partial credit BLEU alone cannot.

\subsection{Classification}
\label{sec:classification}
Using thresholds $\tau_{\text{text}}{=}0.4$, $\tau_{\text{low}}{=}0.5$, and
$\tau_{\text{high}}{=}0.8$, with $b{=}\mathrm{BLEU}(r,g)$ and
$c{=}\mathrm{cos}(r,g)$:
\begin{equation}
  \mathrm{label}(b,c) =
  \begin{cases}
    \mathrm{hallucinated} & \text{if } c < \tau_{\text{low}}, \\
    \mathrm{correct}
      & \text{if } c \ge \tau_{\text{low}},\;
        b > \tau_{\text{text}},\;
        c > \tau_{\text{high}}, \\
    \mathrm{theme\_only}
      & \text{if } c \ge \tau_{\text{low}},\;
        b \le \tau_{\text{text}},\;
        c > \tau_{\text{high}}, \\
    \mathrm{partial} & \text{otherwise.}
  \end{cases}
  \label{eq:classify}
\end{equation}
Intuitively, \emph{hallucinated} fails the semantic floor outright,
\emph{correct} clears both thresholds, \emph{theme\_only} is on-theme but
wrongly worded, and \emph{partial} covers the remaining borderline cases.
Thresholds were fixed once from pilot inspection and held constant across
all models.

% ─────────────────────────────────────────────────────────────────────────────
\section{Results}
\label{sec:results}

Table~\ref{tab:main-results} reports aggregate scores across all six
collections per model per language (30,000 responses total). The \emph{correct}
rate is 0.0\% and the refusal rate is negligible ($\le$0.7\%) in every cell;
both columns are omitted from the table.

% ── Main results table ────────────────────────────────────────────────────────
\begin{table}[t]
  \centering
  \caption{Hadith-recall results aggregated over all six collections
    (3,000 responses per model--language cell). Correct\,=\,0.0\% and
    Refused\,$\le$\,0.7\% in every row; remaining columns sum to~100\%.}
  \label{tab:main-results}
  \small
  \setlength{\tabcolsep}{6pt}
  \begin{tabular}{llccccc}
    \toprule
    \textbf{Model} & \textbf{Lang}
      & \textbf{Avg.\ BLEU}
      & \textbf{Avg.\ Cosine}
      & \textbf{\%Theme-only}
      & \textbf{\%Partial}
      & \textbf{\%Halluc.} \\
    \midrule
    Mistral-7B     & English & 0.007 & 0.559 &  0.47 & 75.03 & 24.50 \\
    Mistral-7B     & Arabic  & 0.006 & 0.719 & 32.27 & 59.73 &  8.00 \\[3pt]
    Gemma3-4B      & English & 0.008 & 0.558 &  0.23 & 74.23 & 25.53 \\
    Gemma3-4B      & Arabic  & 0.019 & 0.736 & 39.83 & 50.80 &  9.37 \\[3pt]
    Phi4-mini-3.8B & English & 0.005 & 0.447 &  0.07 & 41.13 & 58.80 \\
    Phi4-mini-3.8B & Arabic  & 0.019 & 0.500 &  6.97 & 44.83 & 47.50 \\[3pt]
    Jais-7B        & English & 0.007 & 0.535 &  0.03 & 66.10 & 33.87 \\
    Jais-7B        & Arabic  & 0.024 & 0.567 &  9.90 & 54.27 & 35.83 \\[3pt]
    ALLaM-7B       & English & 0.006 & 0.518 &  0.27 & 59.30 & 40.43 \\
    ALLaM-7B       & Arabic  & 0.015 & 0.480 &  4.37 & 41.37 & 54.27 \\
    \bottomrule
  \end{tabular}
\end{table}

\subsection{No Model Achieves Reliable Exact Recall}
Across every model and both languages, the \emph{correct} category is empty.
This holds even for the strongest general-purpose models (Mistral-7B,
Gemma3-4B) and for the Arabic-specialized models (Jais-7B, ALLaM-7B),
indicating that Arabic-domain pretraining does not translate into reliable
verbatim Hadith recall. This pattern is visible directly in
Figures~\ref{fig:mistral}--\ref{fig:allam}, where the \%Correct bar remains
at zero for every collection and every model, complementing the aggregate
view in Table~\ref{tab:main-results}.

\subsection{Arabic Lexical Fidelity Is Near-Zero}
Average BLEU for Arabic responses (0.006--0.024) is comparable to or lower
than English BLEU across all models, confirming that no model reproduces
the exact Arabic wording of the requested Hadith. This near-zero fidelity
holds uniformly across every one of the six collections, as the
per-collection BLEU values in Figures~\ref{fig:mistral}--\ref{fig:allam}
show, rather than being concentrated in a small subset of collections.

\subsection{Refusals Are Rare}
Explicit refusals are rare across models, meaning most failures are silent
fabrications rather than abstentions arguably the more dangerous failure mode,
since a user has no explicit signal that the model does not actually know the
answer. Where refusals do appear in the per-collection figures, they are
isolated, nonzero \%Refused values concentrated almost entirely in Arabic
responses from a single model, rather than a behavior shared across the
benchmark.

\subsection{Per-Collection Breakdown}
\label{sec:per-book}

\begin{figure}[t]
  \centering
  \includegraphics[width=\linewidth]{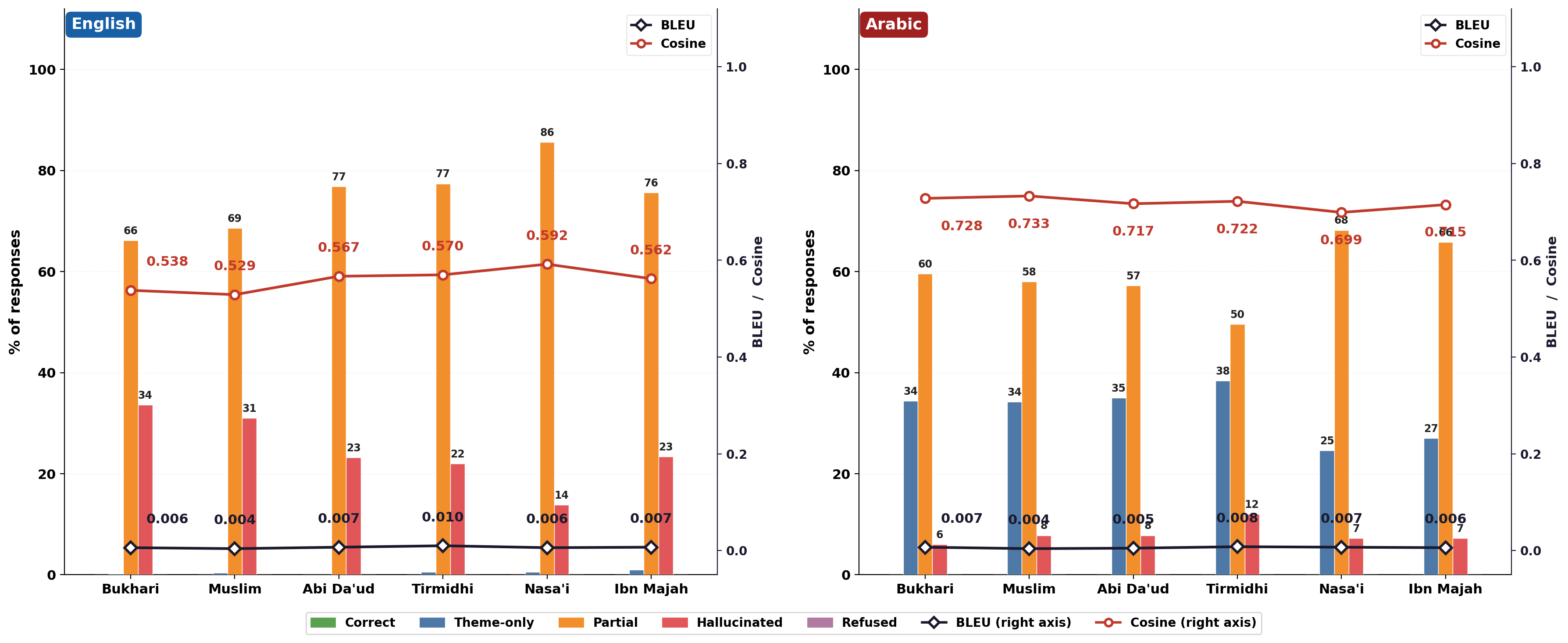}
  \caption{Mistral-7B per-collection results showing \%Correct, \%Theme-only, \%Partial, \%Hallucinated, \%Refused, BLEU, and Cosine similarity for English and Arabic across all six Hadith collections. \%Correct and \%Refused are 0.00\% for both English and Arabic across all collections. }
  \label{fig:mistral}
\end{figure}

\begin{figure}[t]
  \centering
  \includegraphics[width=\linewidth]{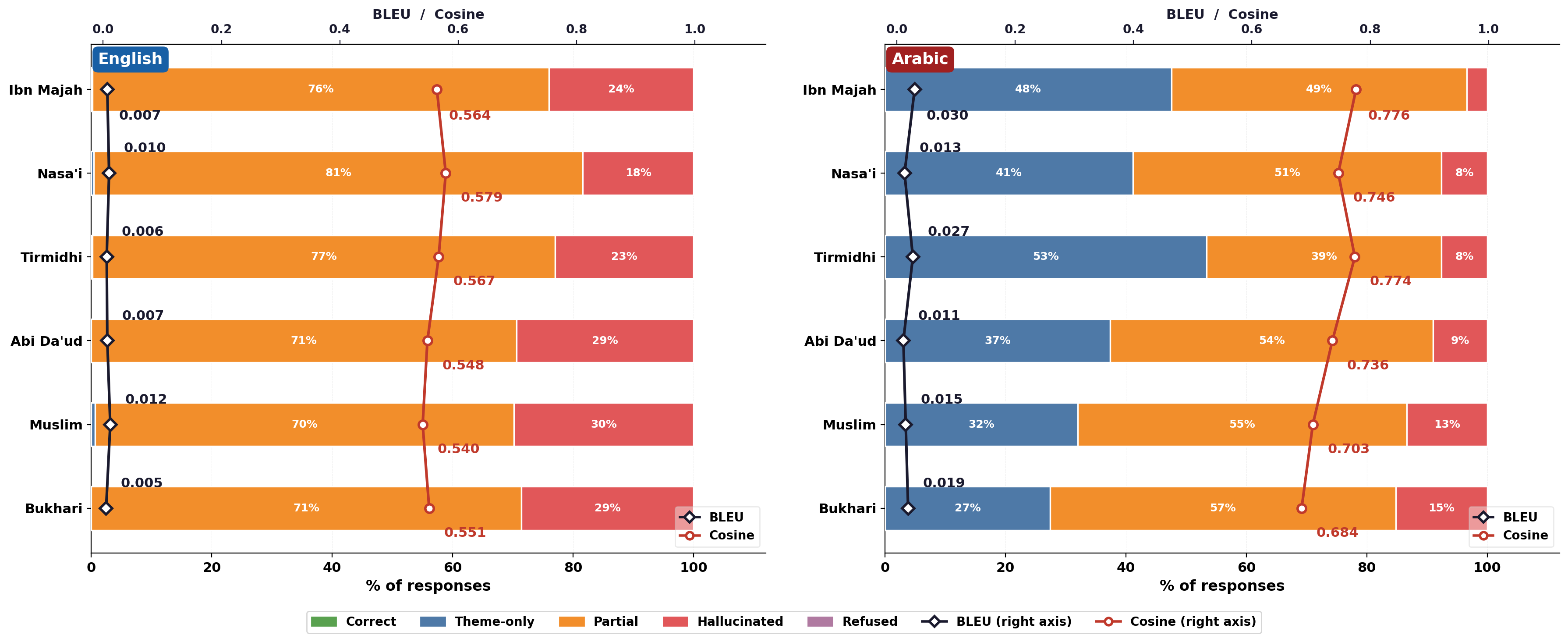}
  \caption{Gemma3-4B per-collection results showing \%Correct, \%Theme-only, \%Partial, \%Hallucinated, \%Refused, BLEU, and Cosine similarity for English and Arabic across all six Hadith collections. \%Correct and \%Refused are 0.00\% for both English and Arabic across all collections.}
  \label{fig:gemma3}
\end{figure}

\begin{figure}[t]
  \centering
  \includegraphics[width=0.90\linewidth]{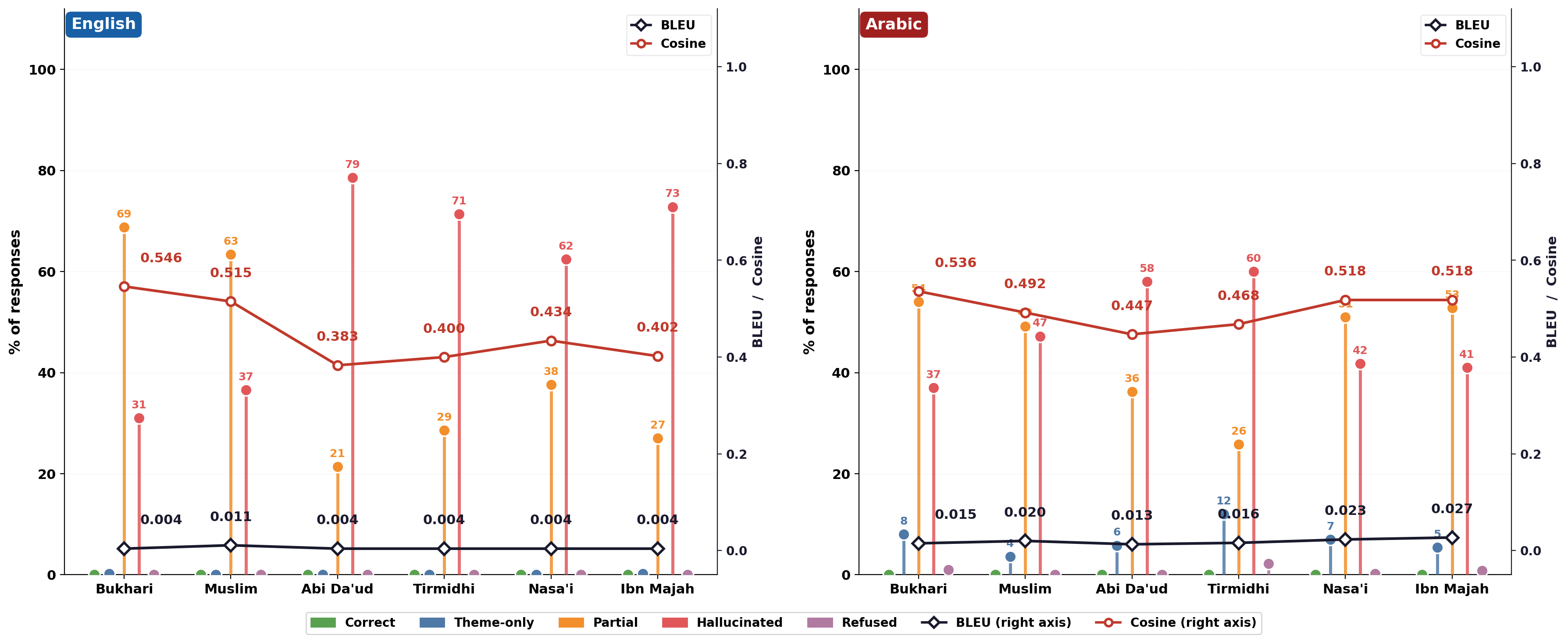}
  \caption{Phi4-mini-3.8B per-collection results showing \%Correct, \%Theme-only, \%Partial, \%Hallucinated, \%Refused, BLEU, and Cosine similarity for English and Arabic across all six Hadith collections. \%Correct is 0.00\% across all collections; Arabic refusals where nonzero: Bukhari 1.0\%, Tirmidhi 2.2\%, Nasa'i 0.2\%, Ibn Majah 0.8\%.}
  \label{fig:phi4mini}
\end{figure}

\begin{figure}[t]
  \centering
  \includegraphics[width=\linewidth]{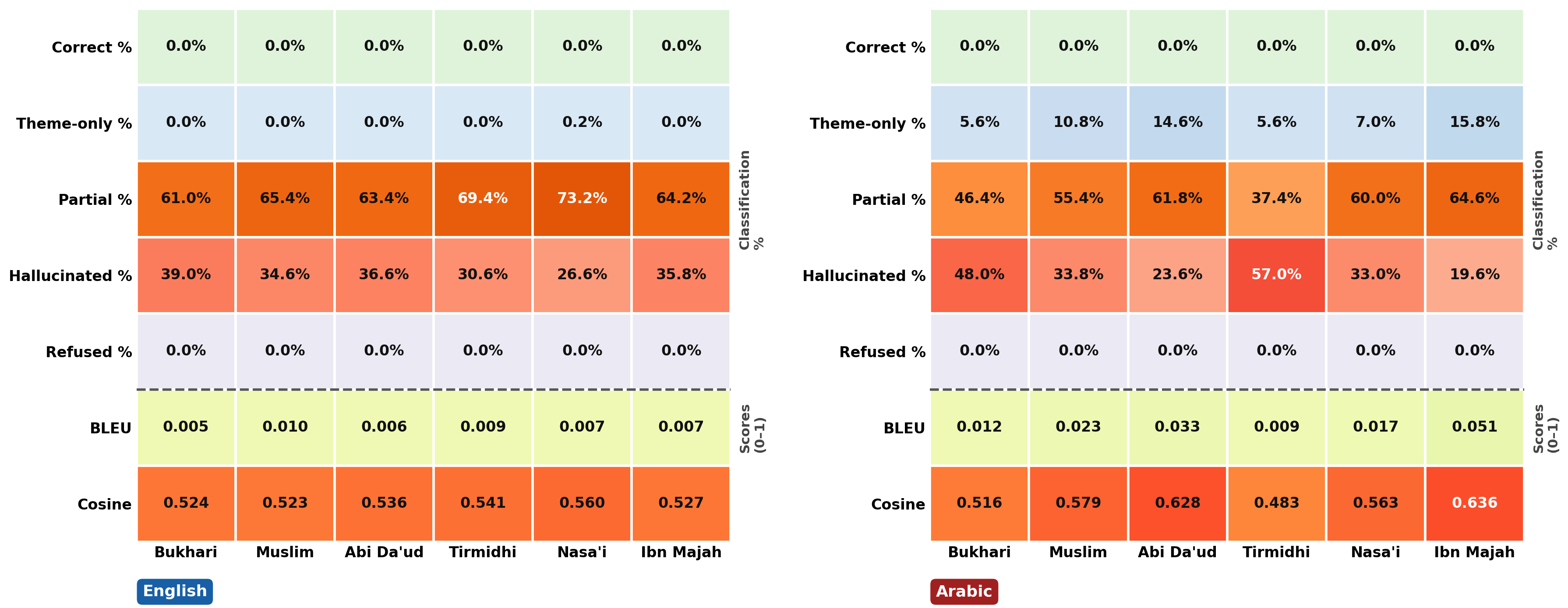}
  \caption{Jais-7B per-collection results showing \%Correct, \%Theme-only, \%Partial, \%Hallucinated, \%Refused, BLEU, and Cosine similarity for English and Arabic across all six Hadith collections. \%Correct and \%Refused are 0.00\% for both English and Arabic across all collections.}
  \label{fig:jais}
\end{figure}

\begin{figure}[t]
  \centering
  \includegraphics[width=\linewidth]{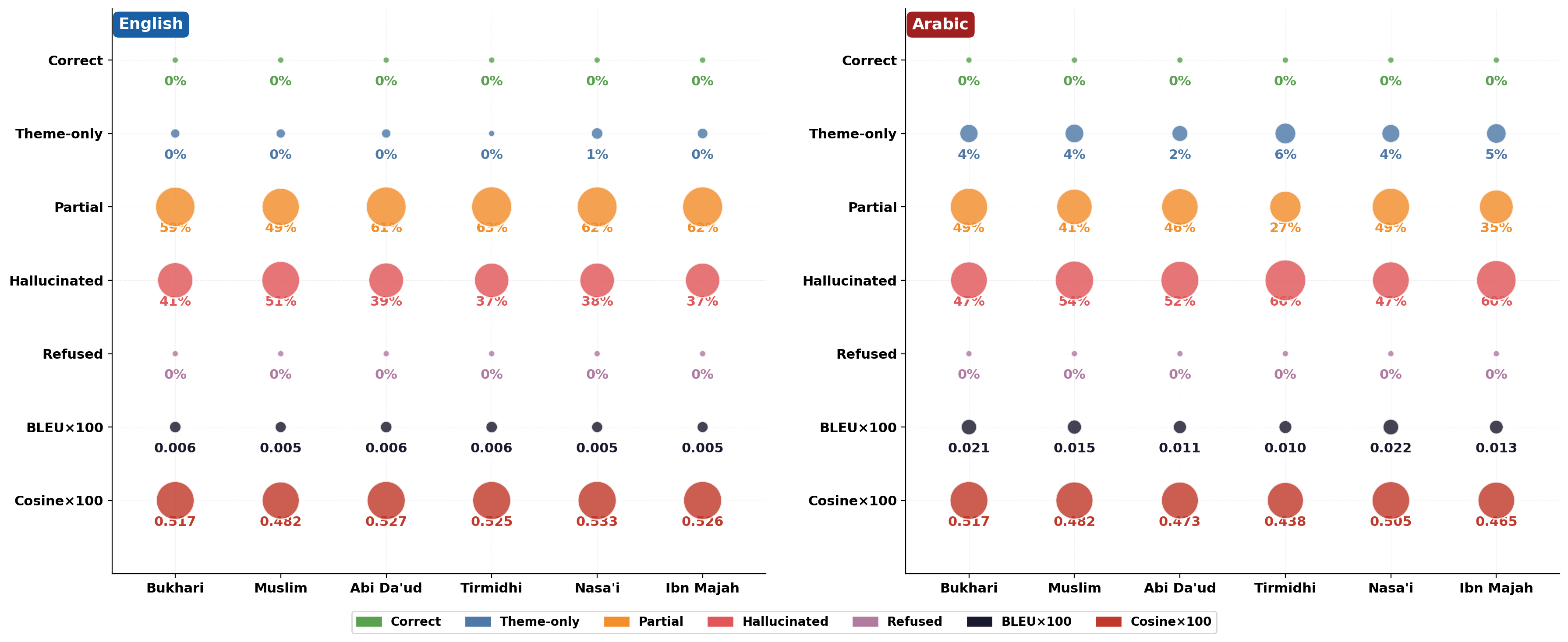}
  \caption{ALLaM-7B per-collection results showing \%Correct, \%Theme-only, \%Partial, \%Hallucinated, \%Refused, BLEU, and Cosine similarity for English and Arabic across all six Hadith collections. \%Correct and \%Refused are 0.00\% for both English and Arabic across all collections.}
  \label{fig:allam}
\end{figure}

Figures~\ref{fig:mistral}--\ref{fig:allam} break the aggregate numbers in
Table~\ref{tab:main-results} down by individual Hadith collection for each
model. Collection abbreviations used throughout: Bukhari, Muslim, Abi Da'ud,
Tirmidhi, Nasa'i, Ibn Majah. Each figure reports \%Correct, \%Theme-only,
\%Partial, \%Hallucinated, and \%Refused alongside average BLEU and cosine
scores, split by language, allowing a direct visual comparison of a given
model's behavior across all six canonical collections.

% Figure 2 – Mistral (floats to top of next available page)

One pattern is consistent across all five per-collection figures: within a
given model and language, BLEU never approaches the correctness threshold
$\tau_{\text{text}}{=}0.4$ for any individual collection the near-zero
lexical fidelity in Table~\ref{tab:main-results} is not an artifact of
averaging over a few well-recalled collections and many poorly-recalled
ones; it holds uniformly, collection by collection (Figure~\ref{fig:mistral}
for Mistral-7B; Figure~\ref{fig:gemma3} for Gemma3-4B).

% Figure 3 – Gemma3

% Figure 4 – Phi4

Phi4-mini-3.8B (Figure~\ref{fig:phi4mini}) shows the highest hallucination
rates in English among the five models, with \%Hallucinated exceeding 70\%
on Abi Da'ud, Tirmidhi, and Ibn Majah collections for which even the
Partial rate is substantially depressed. Arabic responses for Phi4-mini-3.8B
are comparatively more balanced across Partial and Hallucinated, though BLEU
remains near zero in both languages.

% Figure 5 – Jais

The Arabic-specialized models tell a similar story.
Jais-7B (Figure~\ref{fig:jais}) shows no advantage in Arabic lexical
fidelity over the general-purpose models despite its Arabic-adapted
pretraining, with BLEU scores uniformly below 0.06 across all collections
in both languages. ALLaM-7B (Figure~\ref{fig:allam}), trained with
substantial Arabic and Islamic-domain data, likewise fails to approach the
correctness threshold in either language, with Arabic \%Hallucinated
exceeding English \%Hallucinated for every collection the reverse of the
pattern one might expect from domain-specific training.

% Figure 6 – ALLaM

% ─────────────────────────────────────────────────────────────────────────────
\section{Discussion}
\label{sec:discussion}

These results indicate that current 4-8B-parameter open-weight LLMs, run locally
with no retrieval augmentation, cannot be trusted to recall exact Hadith text
from a bibliographic reference in either Arabic or English. This is not a
marginal miscalibration it is a near-total absence of verified recall across
30,000 trials. Two implications follow. First, any user-facing application that
lets people ask an LLM to ``quote'' a Hadith by reference, without grounding
the model in a retrieval pipeline over a verified corpus, is at serious risk of
surfacing confident fabrications that read as authentic religious text. Second,
Arabic-specialized pretraining, at least at the model scales tested here, does
not close this gap suggesting the failure is one of \emph{memorization of a
long-tail, exact-text corpus}, not general language competence in Arabic.

This benchmark is presented as a call to action for the community building
LLM tools for Muslim users: prioritize retrieval-augmented designs with a
verified Hadith database as the source of truth, add refusal/uncertainty-signaling
for the many silent-fabrication cases observed, and treat verbatim scriptural
recall as a distinct, high-stakes evaluation axis.

% ─────────────────────────────────────────────────────────────────────────────
\section{Limitations}
\label{sec:limitations}

This study uses a publicly available, open-source Hadith dataset, and
independently verified a substantial number of its entries for structural
consistency (row-level completeness, correct book and chapter attribution)
prior to sampling. While its creator has flagged it as a preliminary
release, this verification found no evidence that residual noise would
inflate model performance, so it does not threaten this study's central
near-zero-recall finding. Thresholds were chosen based on manual inspection
and are not learned; the qualitative finding is robust to reasonable
threshold variation given how far below threshold observed BLEU scores
fall. Due to limited hardware resources, this study evaluates five
open-weight models at consumer-hardware scale (3.8--8B parameters) and does
not evaluate larger closed-source models. Finally, this study evaluates
zero-shot, retrieval-free recall by design; how much retrieval augmentation
would close the gap is left to future work.

% ─────────────────────────────────────────────────────────────────────────────
\section{Ethical Considerations}
\label{sec:ethics}

This work studies factual fidelity in the generation of religious scripture and
is intended to surface a safety concern, not to produce or promote fabricated
religious content. All fabricated Hadith text produced during this study is
used strictly for scoring and analysis and is not published, redistributed, or
presented as authentic. Transparently documenting this failure mode is
believed to be in the interest of both the ML and Muslim communities, and
downstream applications are encouraged to adopt retrieval-grounded designs
as discussed in Section~\ref{sec:discussion}.

\begin{ack}
Acknowledgments to be added in the camera-ready version.
\end{ack}

% ─────────────────────────────────────────────────────────────────────────────

% ─────────────────────────────────────────────────────────────────────────────
% \appendix
% \section{Technical Appendix}

% Per-collection breakdowns are reported in
% Figures~\ref{fig:mistral}--\ref{fig:allam} in Section~\ref{sec:per-book}. Full
% response-level scored CSVs (model output, ground truth, BLEU score, cosine
% score, and label for every one of the 30,000 responses) are available in the
% project repository's \texttt{results/} directory and will be released alongside
% the paper.


\begin{thebibliography}{99}
\small

\bibitem{jais} Sengupta, N.\ et al.\ (2023). Jais and Jais-chat: Arabic-centric
foundation and instruction-tuned open generative large language models.
\emph{arXiv preprint}.
\url{https://doi.org/10.48550/arXiv.2308.16149}

\bibitem{allam} Bari, M.\,S.\ et al.\ (2024). ALLaM: Large language models for Arabic
and English. \emph{arXiv preprint}.
\url{https://doi.org/10.48550/arXiv.2407.15390}

\bibitem{hadithdataset} Fahd (2018). \emph{Hadith Dataset}. Kaggle. CC0: Public Domain.
\url{https://www.kaggle.com/datasets/fahd09/hadith-dataset}.

\bibitem{bukhari} al-Bukhari, M.\,I. \emph{Sahih al-Bukhari}. 9th century CE.

\bibitem{muslim} Muslim ibn al-Hajjaj. \emph{Sahih Muslim}. 9th century CE.

\bibitem{abudawud} Abu Dawud, S. \emph{Sunan Abi Da'ud}. 9th century CE.

\bibitem{tirmidhi} al-Tirmidhi, M. \emph{Jami' al-Tirmidhi}. 9th century CE.

\bibitem{nasai} al-Nasa'i, A. \emph{Sunan an-Nasa'i}. 9th century CE.

\bibitem{ibnmajah} Ibn Majah, M. \emph{Sunan Ibn Majah}. 9th century CE.

\bibitem{ollama} Ollama. \emph{Get up and running with large language models locally}.
\url{https://ollama.com}.

\bibitem{mistral} Jiang, A.\,Q.\ et al.\ (2023). Mistral 7B. \emph{arXiv preprint}.
\url{https://doi.org/10.48550/arXiv.2310.06825}

\bibitem{gemma3} Gemma Team et al.\ (2025). Gemma 3 technical report. \emph{arXiv
preprint}. \url{https://doi.org/10.48550/arXiv.2503.19786}

\bibitem{phi4mini} Abouelenin, A.\ et al.\ (2025). Phi-4-Mini technical report.
\emph{arXiv preprint}. \url{https://doi.org/10.48550/arXiv.2503.01743}

\bibitem{bleu} Papineni, K., Roukos, S., Ward, T., \& Zhu, W.\,J. (2002). BLEU: A method
for automatic evaluation of machine translation. \emph{ACL}, pp.\,311--318.
\url{https://doi.org/10.3115/1073083.1073135}

\bibitem{sbert} Reimers, N.\ \& Gurevych, I. (2019). Sentence-BERT: Sentence embeddings
using Siamese BERT-networks. \emph{EMNLP-IJCNLP}, pp.\,3982--3992.
\url{https://doi.org/10.18653/v1/D19-1410}

\end{thebibliography}
\end{document}